\documentclass{article} % For LaTeX2e
\usepackage{iclr2021_conference,times}

\usepackage{amsmath,amsfonts,bm}

\def\eqref#1{equation~\ref{#1}}
\def\1{\bm{1}}

\DeclareMathAlphabet{\mathsfit}{\encodingdefault}{\sfdefault}{m}{sl}
\SetMathAlphabet{\mathsfit}{bold}{\encodingdefault}{\sfdefault}{bx}{n}

\usepackage{hyperref}
\usepackage{url}
\usepackage{booktabs}
\usepackage{tipa}
\usepackage{amssymb}
\usepackage{amsmath}
\usepackage{amsfonts}
\usepackage{enumitem}
\usepackage{caption}
\usepackage{subcaption}
\usepackage{graphicx}
\usepackage{xspace}
\usepackage{float}
\usepackage{wrapfig}

\title{Long Range Arena: A Benchmark for Efficient Transformers}

\author{Yi Tay$^1$\thanks{First two authors contributed equally.} , Mostafa Dehghani$^{1*}$, Samira Abnar$^1$,  Yikang Shen$^1$, Dara Bahri$^1$, Philip Pham$^1$ \\ \textbf{Jinfeng Rao$^1$, Liu Yang$^1$, Sebastian Ruder$^2$, Donald Metzler$^1$} \\
$^1$Google Research \\
$^2$Google DeepMind \\
\texttt{\{yitay, dehghani\}@google.com}

}

\newcommand{\lra}{Long-Range Arena\xspace}

\iclrfinalcopy % Uncomment for camera-ready version, but NOT for submission.
\begin{document}

\maketitle

\begin{abstract}
Transformers do not scale very well to long sequence lengths largely because of quadratic self-attention complexity. In the recent months, a wide spectrum of efficient, fast Transformers have been proposed to tackle this problem, more often than not claiming superior or comparable model quality to vanilla Transformer models. To this date, there is no well-established consensus on how to evaluate this class of models. Moreover, inconsistent benchmarking on a wide spectrum of tasks and datasets makes it difficult to assess relative model quality amongst many models. This paper proposes a systematic and unified benchmark, \emph{\lra}, specifically focused on evaluating model quality under long-context scenarios. Our benchmark is a suite of tasks consisting of sequences ranging from $1K$ to $16K$ tokens, encompassing a wide range of data types and modalities such as text, natural, synthetic images, and mathematical expressions requiring similarity, structural, and visual-spatial reasoning. We systematically evaluate ten well-established long-range Transformer models (Reformers, Linformers, Linear Transformers, Sinkhorn Transformers, Performers, Synthesizers, Sparse Transformers, and Longformers) on our newly proposed benchmark suite. \lra paves the way towards better understanding this class of efficient Transformer models, facilitates more research in this direction, and presents new challenging tasks to tackle. Our benchmark code will be released at \url{https://github.com/google-research/long-range-arena}.
\end{abstract}

\section{Introduction}
Transformers~\citep{vaswani2017attention} are ubiquitously state-of-the-art across many modalities, from language~\citep{devlin2018bert,raffel2019exploring,child2019generating} to images~\citep{Tan2019lxmert,lu2019vilbert} to protein sequences~\citep{rives2019biological}. A common weakness of Transformers is their quadratic memory complexity within the self-attention mechanism that restricts their potential application to domains requiring longer sequence lengths. To date, a dizzying number of efficient Transformer models (\textit{`xformers'}) have been proposed to tackle this problem~\citep{liu2018generating,kitaev2020reformer,wang2020linformer,tay2020sparse,katharopoulos2020transformers}. Many of these models demonstrate comparable performance to the vanilla Transformer model while successfully reducing the memory complexity of the self-attention mechanism. An overview of this research area can be found in~\citep{tay2020efficient}.

Comparing the evaluation and experimental setup of many of these papers, we can make the following observations. Firstly, there is no unifying consensus on what makes an acceptable test bed for benchmarking efficient Transformers. There is also a large diversity in the types of tasks adopted---every single model is evaluated on a different set of tasks and datasets, which makes comparison of different models as well as an assessment of their relative strengths and weaknesses difficult. Secondly, the benchmarks used for evaluation are often arbitrarily chosen, without much consideration to whether the task is suitable for evaluating long-range modeling. Thirdly, many papers tend to conflate the effectiveness of the inductive bias with the benefits of pretraining~\citep{ainslie2020etc,zaheer2020big,wang2020linformer}, which tends to obfuscate the true value of the architecture. Pretraining itself is a computationally expensive endeavour and de-coupling inductive bias research from pretraining would make xformer research more accessible. 

% Pertaining to existing benchmarks, language modeling (LM) is a notable example. While it is intuitive that LM relies substantially on nearby context to predict the next word, it is still the most commonly evaluated benchmark in the efficient Transformer repertoire~\citep{child2019generating,kitaev2020reformer,tay2020sparse,beltagy2020longformer}---even if explicit long-range information is not the key focus. Performance evaluation on LM is also nearing its ceiling and saturation point where all efficient Transformers often get very similar scores. 

In this paper, we propose a new benchmark, \textit{\lra} (LRA), for the purpose of benchmarking sequence models under the long-context scenario. We design a benchmark suite comprised of both synthetic probing tasks and real-world tasks and provide relative comparisons for \textbf{ten} recently proposed efficient Transformer models including Sparse Transformers~\citep{child2019generating}, Reformer~\citep{kitaev2020reformer}, Linformer~\citep{wang2020linformer}, Longformer~\citep{beltagy2020longformer}, Sinkhorn Transformers~\citep{tay2020sparse}, Performers~\citep{choromanski2020rethinking}, Synthesizers~\citep{tay2020synthesizer}, Linear Transformers~\citep{katharopoulos2020transformers}, and BigBird~\citep{zaheer2020big}. This is the most comprehensive and extensives side-by-side evaluation of this class of models. 

While the focus of this benchmark is the ability of these architectures to reason in long-context scenarios, we are also fundamentally interested in understanding the capabilities and properties of these xformer architectures when exposed to different types of data and conditions. Hence, our benchmark is purposefully designed to be capability probing, i.e, we select datasets and tasks with certain innate structure. For example, can these architectures model long sequences that are intrinsically hierarchical or that contain some form of spatial structure? In general, we are especially interested in the relative performance of these xformer models across diverse circumstances. We hope that understanding these better will inspire research on more efficient architectures in the future. While the focus of this paper is on efficient Transformer models, our benchmark is also model agnostic and can also serve as a benchmark for long-range sequence modeling.

Aside from comparing the quality of these models, we also conduct extensive efficiency and memory usage analysis of these models. We believe such a side-by-side performance benchmark will be valuable to the community, providing deeper insight on the practical efficiency of these methods. Overall, we propose a unified framework for enabling easy side-by-side comparisons of efficient Transformer models and broadly speaking, long-range sequence models in general. Our framework, which we open source, is written in JAX/FLAX\footnote{\url{https://github.com/google/flax}}.

\section{\lra (LRA)}
This section introduces the \lra (LRA) benchmark (pronounced \textit{el-ra}). We implement our benchmark (which includes the task, evaluators, and models) in Python 3 and Jax/Flax and open-source our code\footnote{\url{https://github.com/google-research/long-range-arena}}---making it easy to extend and to build on top of our work.

\subsection{Desiderata}
For creating the \lra benchmark, we established a set of desiderata:
\begin{enumerate}[leftmargin=0.6cm]
\item \textbf{Generality}: All efficient Transformers models should be applicable to our tasks. For instance, given that not all xformer models are able to perform autoregressive decoding~\citep{wang2020linformer}, we include tasks that only require encoding. 
\item \textbf{Simplicity}: The tasks should have a simple setup. All factors that make comparisons difficult should be removed. This encourages simple models instead of cumbersome pipelined approaches. For instance, we avoid including any  particular data augmentation and consider pretraining to be out of scope of this benchmark. 
\item \textbf{Challenging}: The tasks should be difficult enough for current models to ensure there is room for improvement to encourage future research in this direction.
\item \textbf{Long inputs}: The input sequence lengths should be reasonably long since assessing how different models capture long-range dependencies is a core focus of LRA.
\item \textbf{Probing diverse aspects}: The set of tasks should assess different capabilities of models like their ability to model relations and hierarchical/spatial structures, generalization capability, etc.
\item \textbf{Non-resource intensive and accessible}: The benchmarks should be deliberately designed to be lightweight so as to be accessible to researchers without industry-grade computing resources.
\end{enumerate}

\subsection{Tasks}

This section describes the tasks in the LRA benchmark. Note that these tasks are specifically designed for the purpose of assessing different aspects of efficient Transformer models. Further details about each task can be found in the appendix.

      \subsubsection{Long ListOps} 
      In this task, we are interested in the capability of modeling hierarchically structured data in a long-context scenario. This benchmark task is a longer variation of the standard ListOps task proposed in~\citep{nangia2018listops}, which was designed to investigate the parsing ability of neural models. 
      
      The dataset is comprised of sequences with a hierarchical structure and operators \textsc{max}, \textsc{mean}, \textsc{median} and \textsc{sum\_mod} that are enclosed by delimiters (brackets). An example (much shorter) sequence is as follows:
      
      \begin{minipage}{\textwidth}
      \centering{
      \fontsize{9}{9}\fontfamily{pcr}\selectfont
       \textbf{INPUT}: [{MAX} 4 3 [{MIN} 2  3  ]  1  0  [{MEDIAN} 1  5  8 9, 2]] 
       ~~~~~ \textbf{OUTPUT}: 5
      }\end{minipage}
      
      In our task we use a version of ListOps of sequence lengths of up to $2K$ to test the ability to reason hierarchically while handling long contexts. Naturally, in the above example the model needs to access all tokens and model the logical structure of the inputs in order to make a prediction. The task is a ten-way classification task and is considerably challenging. 
      
    \subsubsection{Byte-level Text Classification} This task using real-world data represents a common use case of efficient Transformers, which are often needed to process long documents. Text classification in particular is associated with many real-world applications such as spam, fraud, and bot detection and commercial document classification, among others \citep{Howard2018}.
    
    This task also benchmarks the ability of the models to deal with compositionality as it is required to compose characters into words into higher-level phrases. Compared to ListOps, boundaries are less well defined and need to be learned from the data, which is a challenging problem in its own right \citep{Kawakami2019learning}.
    
    We consider the byte/character-level setup of this task in order to simulate a longer input sequence, which also makes the task considerably more challenging.\footnote{On the IMDb word-level task, models without pre-training achieve accuracies in the high 80s while the same models score in the mid 60s on the character-level task \citep{tay2020sparse}.} Note that this setup differs significantly from character-level language modeling (char LM). In char LM, it would suffice to read nearby context to determine the next character, e.g., a model is very likely to predict \textit{`e'} after having seen the prefix \textit{`appl'}. For byte-level text classification, the model needs to reason with compositional, unsegmented data in order to solve a meaningful real-world task. We use the IMDb reviews~\citep{maas-EtAl:2011:ACL-HLT2011} dataset, which is a commonly used dataset to benchmark document classification. We use a fixed max length of $4K$ for this task, which is truncated or padded when necessary. This is a binary classification task with accuracy as the metric.
    \subsubsection{Byte-level Document Retrieval}
    We further evaluate a model's ability to encode and store compressed representations that are useful for matching and retrieval. Learning the similarity score between two vectors is a common problem in machine learning and is useful for a wide array of applications \citep{guo2016deep}. 
    
    Hence, this task is mainly about modeling a similarity score between two documents in a `two tower setup' in which compressed representations are concatenated and passed into a linear classifier. Note that we deliberately prevent models from using cross attention. This task thus serves as a test of how well models are able to \textit{compress} long sequences into representations suitable for similarity-based matching. 
    
    We use the ACL Anthology Network \citep[AAN;][]{radev2013acl} dataset, which identifies if two papers have a citation link, a common setup used in long-form document matching~\citep{jiang2019semantic,DBLP:journals/corr/abs-2004-12297}. 
    
    Similar to the text classification setup, we use a byte/character level setup, which challenges the model to compose and aggregate information over longer contexts. We use a sequence length of $4K$ for each document, which makes the total text length $8K$ for this task. This is a binary classification task with accuracy as the metric.
    \subsubsection{Image Classification on sequences of pixels} 
    This task is an image classification task, where the inputs are sequences of pixels. In other words, an $N \times N$ image is flattened to a sequence of length $N^2$ pixels. 
    
    Similar to how the previous tasks require capturing the hierarchical structure in the data, this task requires the model to learn the 2D spatial relations between input pixels, while presented as a 1D sequence of symbols. 
    
    We focus on assessing Transformer models that are designed to process a sequence of discrete symbols, so we do not allow extra modules such as a CNN stem that embeds pixel-level inputs. To simplify the setup, we map the input images to a single gray-scale channel where each pixel is represented with an 8-bit pixel intensity (vocabulary size of 256). In LRA, we use the CIFAR-10 dataset~\citep{Krizhevsky09learningmultiple} for the image classification task.
    
    \subsubsection{Pathfinder (Long-Range Spatial Dependency)} 
    \begin{wrapfigure}{t}{0.38\textwidth}
    \vspace{-10pt}
     \centering
     \begin{subfigure}[b]{0.3\textwidth}
         \centering
         \includegraphics[width=\textwidth]{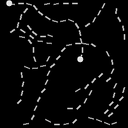}
         \vspace{-5pt}
         \caption{A positive example.}
         \label{fig:path_finder_pos}
     \end{subfigure}
    % \hspace{20pt}
     \begin{subfigure}[b]{0.3\textwidth}
         \centering
         \includegraphics[width=\textwidth]{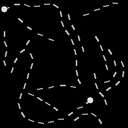}
         \vspace{-5pt}
         \caption{A negative example.}
         \label{fig:path_finder_neg}
     \end{subfigure}
     \caption{Samples of the Pathfinder task.}
    \label{fig:path_finder}
    \vspace{-45pt}
    \end{wrapfigure}

    The Pathfinder challenge~\citep{linsley2018learning, Kim2020Disentangling} was first introduced for learning long-range spatial dependencies. It is a synthetic visual task motivated by cognitive psychology~\citep{houtkamp2010parallel}. 
    
    The task requires a model to make a binary decision whether two points represented as circles are connected by a path consisting of dashes. We show a positive example of two connected points and a negative example of two unconnected points in Figure \ref{fig:path_finder}. The dataset also contains distractor paths, which makes this setup challenging. We model this task by treating images as sequences of pixels. In this task, images are of dimensions ($32 \times 32)$, which make up a sequence length of $1024$. 

    \subsubsection{Pathfinder-X (Long-Range Spatial Dependencies with Extreme Lengths)}
    Finally, we consider an extreme version of Pathfinder (Pathfinder-X) where examples consist of $16K$ pixels (i.e., images of $128 \times 128$). 
    
    The key goal here is to observe if a model would fail to solve the $16K$ extreme version even if it can successfully learn the standard version of $1024$ tokens. This is an interesting litmus test to see if the same algorithmic challenges bear a different extent of difficulty when sequence lengths are much longer. We include this in our benchmark as Path-X.
    
\subsection{Required Attention Span of LRA tasks}
One of the main goals of the LRA benchmark is assessing the ability of different efficient Transformer models to capture long-range dependencies. The tasks and setups are designed with this goal in mind. In order to have a quantitative estimate of the spatial extent needed to be considered by an attention mechanism to encode the inputs, we define \emph{required attention span}. 

Given a trained attention-based model and a sequence of tokens as inputs, the required attention span of an attention module is computed as the mean distance between the query token and the attended tokens, scaled by attention weights. Here, we compute the mean \emph{required attention span} over all attention modules in our best vanilla Transformer model for each task, averaged over 1K random samples from the validation set. 

\begin{figure}[H]
\centering
 \includegraphics[width=0.4\textwidth]{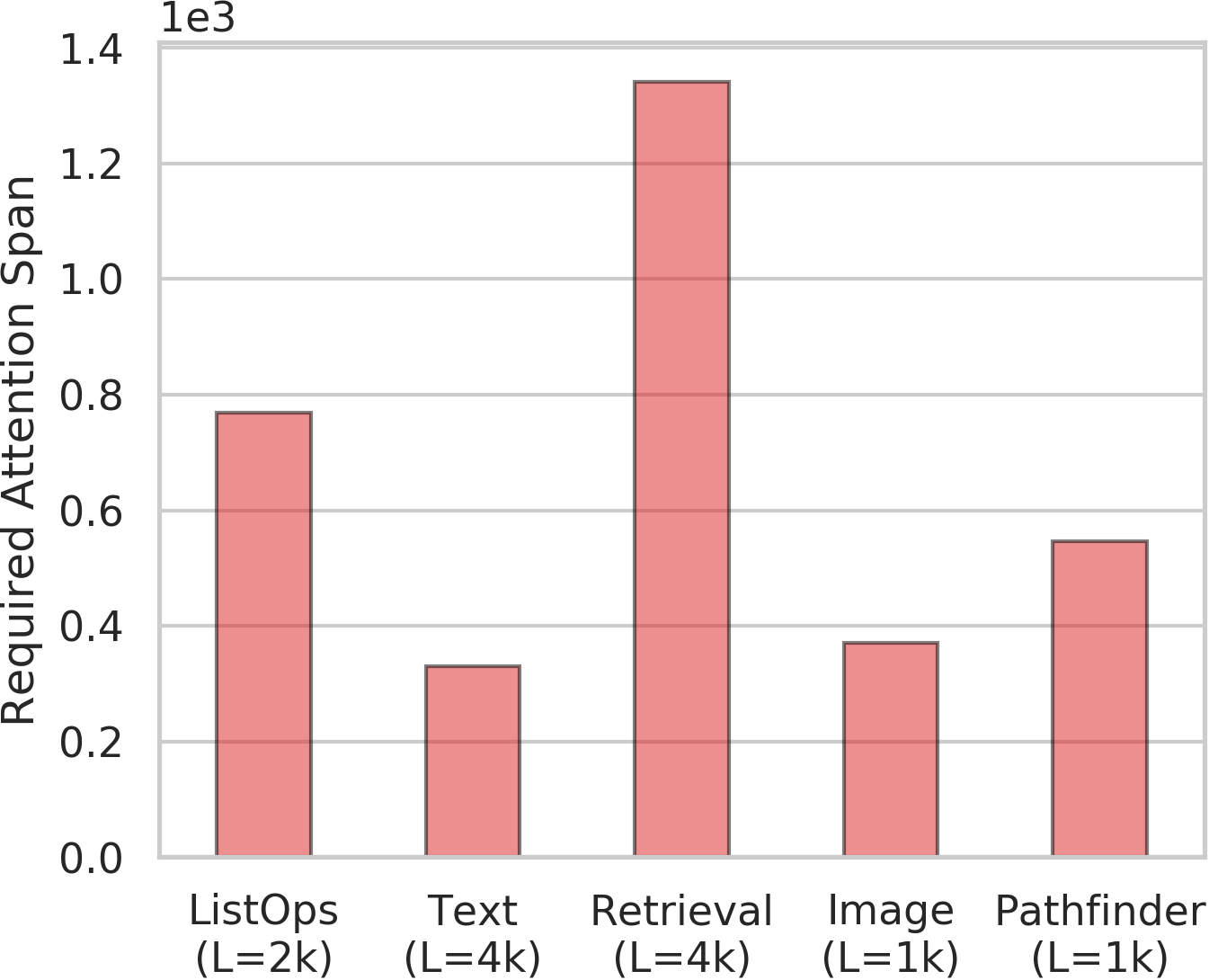}
    \caption{Required attention span on different tasks.}
    \label{fig:req_att_span}
\end{figure}
Figure~\ref{fig:req_att_span} summarizes the required attention span for each task in LRA. 
For all the tasks in LRA the required attention span is rather high. This shows, a Transformer model needs to go beyond combining only local information, while in many other tasks and datasets, attention mechanism mostly need to combine information from neighboring positions.

Given the purpose of LRA, we found \emph{required attention span} serves as a good proxy for how difficult a task is for Transformer-based models.\footnote{Note that this metric mainly provides an indication of the required attention span for a task and the relative differences between tasks based on a reasonably strong model; a better model might only need to attend to shorter ranges \citep{Daniluk2017,Rae2020}.}

\begin{table}[]
    \centering
    \begin{tabular}{l|cccccc|c}
    \toprule
      Model    & ListOps & Text  & Retrieval &  Image & Pathfinder & Path-X & Avg \\
   \midrule
        Transformer &  36.37 & 64.27 &
        57.46 &  42.44& 71.40 & FAIL & \underline{54.39}\\
           \midrule
        Local Attention &  15.82 &52.98 & 53.39 & 41.46& 66.63 & FAIL & 46.06 \\
        Sparse Trans.  & 17.07 & 63.58 & \textbf{59.59} & \textbf{44.24} & 71.71 & FAIL  & 51.24 \\
        Longformer& 35.63& 62.85 & 56.89&  42.22 & 69.71 & FAIL & 53.46 \\
        Linformer &   35.70 & 53.94&  52.27 & 38.56 & \underline{76.34} & FAIL & 51.36\\
        Reformer &  \textbf{37.27} & 56.10 & 53.40 &  38.07& 68.50 & FAIL & 50.67 \\
        Sinkhorn Trans. &33.67 & 61.20 & 53.83 & 41.23 & 67.45 & FAIL & 51.39\\
        Synthesizer & \underline{36.99} & 61.68 & 54.67  &41.61 & 69.45 & FAIL & 52.88\\
        BigBird  & 36.05 & 64.02 & \underline{59.29}  &  40.83 & 74.87 & FAIL & \textbf{55.01} \\
        Linear Trans. & 16.13&  \textbf{65.90} & 53.09 & 42.34 & 75.30 & FAIL & 50.55 \\ 
        Performer  &18.01& \underline{65.40} & 53.82 & \underline{42.77} & \textbf{77.05} & FAIL & 51.41\\
        \midrule 
        Task Avg (Std) & 29 (9.7) & 61 (4.6) & 55 (2.6) & 41 (1.8) & 72 (3.7) & FAIL & 52 (2.4)\\ 
      \bottomrule
    \end{tabular}
    \caption{Experimental results on \lra benchmark. Best model is in boldface and second best is underlined. All models do not learn anything on Path-X task, contrary to the Pathfinder task and this is denoted by FAIL. This shows that increasing the sequence length can cause seriously difficulties for model training. We leave Path-X on this benchmark for future challengers but do not include it on the Average score as it has no impact on relative performance.}
    \label{tab:my_label}
\end{table}

\section{Experimental Results}
\label{sec:results}

\subsection{Models}
This section describes the models we evaluate on our LRA benchmark. We base our evaluation on ten recently proposed efficient Transformer models. Aside from the standard vanilla Transformer~\citep{vaswani2017attention} and a simple local attention baseline, we compare Sparse Transformers~\citep{child2019generating}, Longformers~\citep{beltagy2020longformer}, Linformers~\citep{wang2020linformer}, Reformers~\citep{kitaev2020reformer}, Sinkhorn Transformers~\citep{tay2020sparse}, Synthesizers~\citep{tay2020synthesizer}, BigBird~\citep{zaheer2020big}, Linear Transformers~\citep{katharopoulos2020transformers}, and Performers~\citep{choromanski2020masked}. 

We believe these ten models to represent a diverse cross-section of recent efficient Transformer models.

\subsection{Philosophy behind the benchmark}
We note that it is non-trivial and almost impossible to conduct a perfectly fair evaluation of all models. The large search space motivates us to follow a set of fixed hyperparameters (number of layers, heads, embedding dimensions, etc.) for all models. The best performance and relative order of the models may change if we aggressively tune hyperparameters for all models. Hence, the results provided in this paper are not meant to be a final authoritative document on which xformer is the best. Instead, we provide a starting point for future research and strive to be as \textbf{fair} as possible. In order to do so, we plan to release the code with all the hyperparameters and implementation details. Additionally, we intend for our paper to be a living document and encourage researchers (authors and the broader community) to contribute and continue updating this paper (with rules and limitations described in the appendix). We also implemented all models to the best of our abilities. We often consulted with the original developers of the included models.

\subsection{Quantitative Results}
Based on our results, we observe that (1) all proposed tasks in LRA are considerably challenging and (2) there are meaningful differences in model performance across different xformer models. 
\paragraph{Results on ListOps} The ListOps task (10-way classification) has proven to be reasonably difficult with the best models obtaining only $37\%$. The considerable gap to random chance shows that models are indeed learning the task. We notice that the inductive bias of the xformer model plays a substantial role on this task in which approximately half the xformer models are able to get $>30\%$ performance while the remainder of the models only get slightly above random chance. This may imply that certain efficiency-inspired inductive biases may be better at handling hierarchical data than others. For instance, the results from our experiments seem to suggest that kernel-based models (e.g., Performer, Linear Transformers) are possibly not as effective on hierarchically structured data. 
\paragraph{Results on Text Classification} Byte-level classification is shown to be difficult and challenging especially when no pretraining or contextual embeddings are used. The best model only obtains $65.90$ accuracy. The Linear Transformer performs well on this task, along with the Performer model. Contrary to the ListOps task, it seems like fast kernel-based models do well on this task.
\paragraph{Results on Retrieval} The scores of different models on this task are also rather low (average of $55\%$), indicating the difficulty of the task. The vanilla Transformer model only achieves $57.46\%$ accuracy with some xformer variants scoring very close to random chance. The best performing model is the Sparse Transformer and the second best is BigBird. We find that models that follow fixed sparse patterns to do well on this task. Models that are based on low-rank factorization and kernels perform relatively worse.

\paragraph{Results on Image Classification} On the image classification task, most models perform quite similarly (low variance amongst model performance). The best model on this task is the Sparse Transformer, followed by the Performer. Linformer and Reformers do not do well on this task. On a related note, we also observed most of models struggle generalizing to the test even though they manage to overfit the training set. While we extensively tried different regularization techniques on every single model, there is a rather large gap between their performance on train and test set (More details in Appendix).

\paragraph{Results on Pathfinder / Path-X} Results show that all models achieve reasonable performance on the Pathfinder task. The average performance is $72$ and the best model Performer obtains $77.05\%$ accuracy. The Local Attention model performs the worse out of all models.

All models failed to solve the Path-X task, achieving at best $50\%$. We find this intriguing because this is essentially an identical task to the standard Pathfinder, albeit with much longer sequence lengths. Hence, we observe that the extreme length of the task can significantly obstruct a model from leaning anything meaningful. We leave Path-X in our benchmark suite, hoping to spur future progress in modeling sequences at extreme lengths. 

\subsection{Efficiency Benchmarks}
In this section, we report efficiency metrics of our runs. For simplicity, we use the byte-level text classification benchmark and report run times and memory consumption of the sequence lengths $\{1K,2K,3K,4K\}$. We use a batch size of $32$ for all runs and conduct experiments on 4x4 TPU V3 Chips. We emphasize that this is again largely conditioned on hardware and implementation details (more details can be found in the appendix).

\begin{table}[t]
    \centering
    \begin{tabular}{l|cccc|cccc}
    \toprule
    & \multicolumn{4}{c|}{Steps per second} & \multicolumn{4}{c|}{Peak Memory Usage (GB)}\\ 
       Model  & 1K & 2K & 3K & 4K & 1K & 2K & 3K & 4K  \\
       \midrule
     Transformer    & 8.1 & 4.9  &2.3 & 1.4 & 0.85 & 2.65 & 5.51 & 9.48\\
     \midrule
     Local Attention & 9.2 ($1.1$x)& 8.4 ($1.7$x)& 7.4 ($3.2$x) & 7.4 ($5.3$x)& 0.42  & 0.76 & 1.06 & 1.37\\
     Linformer & \underline{9.3} ($1.2$x) &  9.1 ($1.9$x)& 8.5  ($3.7$x)&  7.7 ($5.5$x)&  \textbf{0.37} & \textbf{0.55} & 0.99 & \textbf{0.99}\\
     Reformer & 4.4 ($0.5$x)&  2.2 ($0.4$x)& 1.5 ($0.7$x) & 1.1 ($0.8$x) & 0.48 & 0.99 & 1.53 & 2.28\\
     Sinkhorn Trans & 9.1 ($1.1$x) & 7.9  ($1.6$x) &6.6 ($2.9$x)& 5.3 ($3.8$x) & 0.47 & 0.83 & 1.13 & 1.48\\ 
     Synthesizer & 8.7 ($1.1$x)& 5.7 ($1.2$x) & 6.6 ($2.9$x) & 1.9 ($1.4$x) & 0.65 & 1.98 & 4.09 & 6.99\\
     BigBird & 7.4 ($0.9$x) & 3.9 ($0.8$x) & 2.7 ($1.2$x) & 1.5 ($1.1$x) & 0.77 & 1.49 & 2.18 & 2.88\\ 
     Linear Trans. & 9.1 ($1.1$x)& \underline{9.3} ($1.9$x)& \underline{8.6} ($3.7$x) & \underline{7.8} ($5.6$x)& \textbf{0.37} & \underline{0.57} & \textbf{0.80} & \underline{1.03}\\ 
     Performer & \textbf{9.5} ($1.2$x) & \textbf{9.4} ($1.9$x) &   \textbf{8.7} ($3.8$x) & \textbf{8.0} ($5.7$x)  & \textbf{0.37}  & 0.59 & \underline{0.82} & 1.06
     \\
     \bottomrule
     
    \end{tabular}
    \caption{Benchmark results of all Xformer models with a consistent batch size of $32$ across all models. We report relative speed increase/decrease in comparison with the vanilla Transformer in brackets besides the steps per second. Memory usage refers to per device memory usage across each TPU device. Benchmarks are run on 4x4 TPU V3 Chips.}
    \label{tab:efficiency}
\end{table}
\paragraph{Results on Speed} Table \ref{tab:efficiency} reports our efficiency benchmarks on the xformer models. We note that low-rank and kernel-based models are generally the fastest. The overall fastest model is the Performer model~\citep{choromanski2020masked}, which is 5.7$\times$ faster than Transformers on the $4k$ sequence length. Linformer~\citep{wang2020linformer} and Linear Transformers~\citep{katharopoulos2020transformers} come in a close second and are almost as fast as Performers (at 5.5 to 5.6$\times$ faster). Based on our implementation, the slowest model is the Reformer model~\citep{kitaev2020reformer} that is about $80\%$ the speed of vanilla Transformer at $4K$ sequence lengths and half the speed at $1K$ sequence length.

\paragraph{Results on Memory Consumption} The model with the smallest memory footprint in our benchmarks is the Linformer model, coming in at 0.99GB per TPU device as compared to 9.48GB per TPU device for the vanilla Transformers at $N=4K$. That is about a $10$x reduction in memory footprint. Similar to speed, Performers and Linear Transformers are also relatively compact and are almost as compact as Linformers. Other models (Local Attention, Reformers, BigBird, Synthesizers) are still less memory hungry compared to vanilla Transformers but are relatively less efficient (memory consumption wise) compared to Linformers, Performers, and Linear Transformers. We also notice that the memory consumption of models such as Linformer and Performer scales very well, with the memory usgae at $3K$ and $4K$ being approximately equal.

\subsection{Overall Results: No One-size-fits-all} 

\begin{figure}[H]
   \centering
    \includegraphics[width=0.6\textwidth]{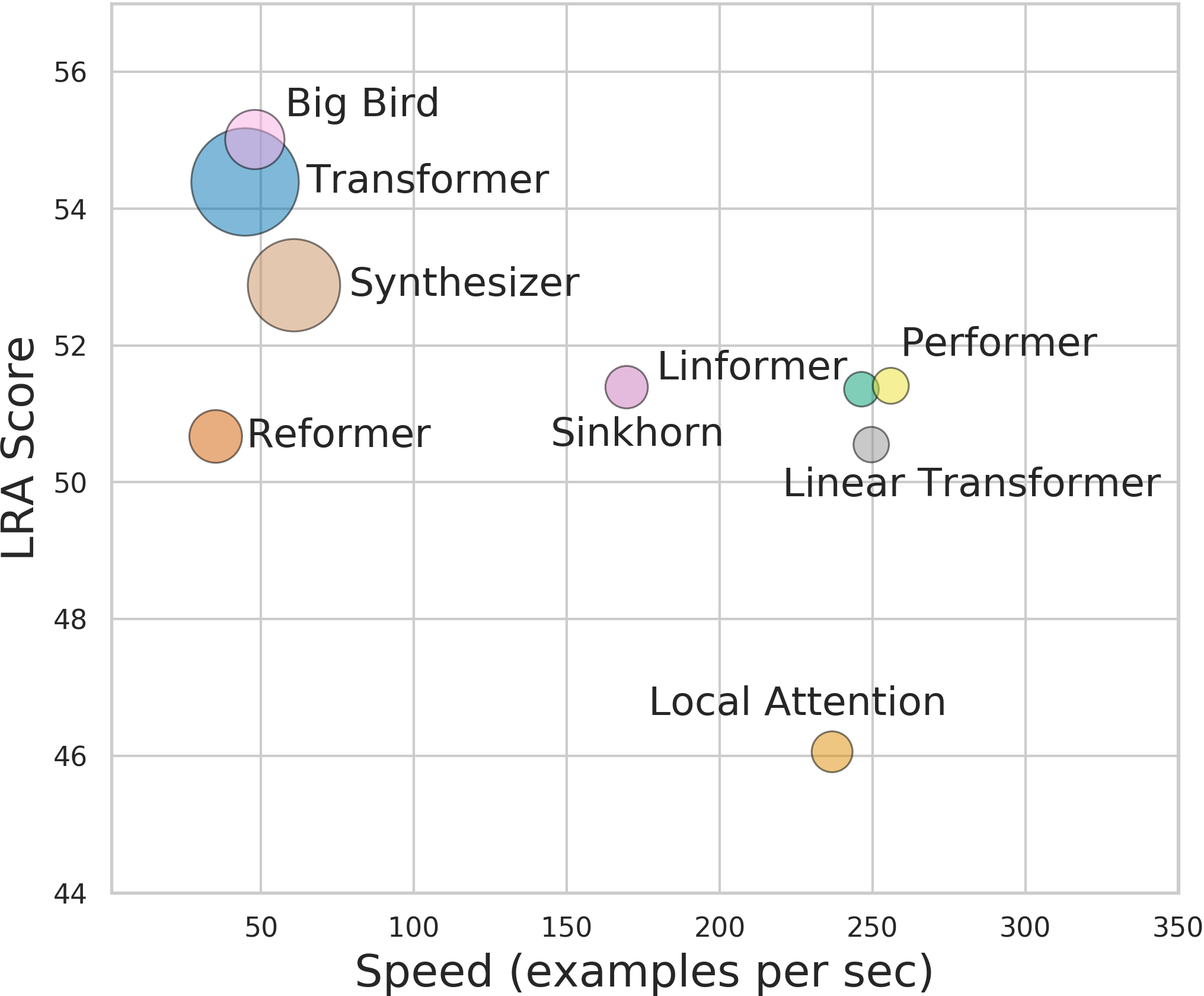}
    \caption{Performance ($y$ axis), speed ($x$ axis), and memory footprint (size of the circles) of different models.}
    \label{fig:perf_spped_mem}
\end{figure}

% \begin{minipage}{0.5\linewidth}
Based on our analysis, the best qualitative performance in terms of LRA score, i.e. integrated across all five tasks, is the BigBird model. While BigBird does not do extremely well on any individual task compared to other models, it has consistently good performance across all tasks. Performers and Linear Transformers have strong performance on some tasks but their average is lowered by the ListOps task. 
Figure~\ref{fig:perf_spped_mem} shows the trade-off between qualitative performance, model speed, and memory footprint. 
While BigBird performs well, its speed is almost similar to the vanilla Transformer. On the other hand, a model like Local Attention is fast at the cost of lower quantitative performance. 
Among these models, the kernel-based variants, i.e., Performer, Linformer, and linear Transformer seem to be able to make a better trade-off in terms of speed and performance, while having reasonable memory usage.
% \end{minipage}
% \begin{minipage}{0.5\linewidth}

% \end{minipage}

\section{Related Work}

\subsection{Efficient Transformers}
The pervasiveness of Transformer models, along with its well-known trait of being memory intensive, has spurred on a large number of innovations on this front. Early work in this area has typically considered a fixed pattern (local window) approach~\citep{liu2018generating,parmar2018image}. More advanced models have been proposed recently, including combined patterns~\citep{child2019generating,ho2019axial,beltagy2020longformer,zaheer2020big}, learned patterns~\citep{kitaev2020reformer,roy2020efficient}, and recent models based on kernels~\citep{katharopoulos2020transformers,choromanski2020masked} or low-rank approximations~\citep{wang2020linformer}. For the sake of brevity, we refer interested readers to~\citep{tay2020efficient} for a detailed survey of this line of research.

\subsection{Existing Benchmarks} 

\paragraph{Generative Modeling / Language Modeling} This generative modeling task requires predicting the next character, word, or pixel and is a staple in xformer evaluations~\citep{roy2020efficient,kitaev2020reformer}. However, it has been debated how much long-range signal such tasks actually encode~\citep{Rae2020}.

LSTM language models augmented with attention have been shown to rarely attend beyond seven preceding words of context~\citep{Daniluk2017} and samples from LSTM language models are known to quickly devolve into generic text. On the other hand, recent models such as the Transformer-XL~\citep{Dai2019transformerxl} have been observed to be sensitive to a context of around 900 tokens and samples from large-scale models~\citep{Radford2019} maintain a consistent theme over much longer sequences. Even such recent models, however, can be improved by limiting the range of attention~\citep{Rae2020}. In sum, while standard language modelling datasets contain \emph{some} long-range signal, which is required to perform long-range coreference resolution, reasoning with events, discourse understanding, etc.~\citep{ruder2019transfer} it seems to be overshadowed by the much stronger signal of short-term word co-occurrences and is thus difficult to evaluate.\footnote{Datasets such as LAMBADA~\citep{Paperno2016lambada} more explicitly test for context understanding but are still restricted to comparatively short contexts of five sentences on average.} 

\paragraph{Question Answering} Another commonly used evaluation task is question answering~\citep[QA;][]{zaheer2020big}. Open-domain QA in particular typically requires the model to answer questions based on long contexts such as entire Wikipedia documents \citep{Joshi2017triviaqa,Kwiatkowski2019naturalquestions} or even books \citep{Kocisky2018narrativeqa}. Other datasets are explicitly designed to require multiple `hops' of reasoning \citep{Welbl2018wikihop,Yang2018hotpotqa}. Successful approaches are often highly engineered, computationally expensive systems that require pre-training and a separate retrieval model \citep{Lee2019latent,Guu2020realm}.

\paragraph{Natural Language Understanding / GLUE tasks} Evaluation on natural language understanding (NLU) tasks is also common~\citep{wang2020linformer}. Examples in most of these datasets such as MultiNLI \citep{Williams2018multinli} and SST \citep{Socher2013sst} consist of single sentences and less than $100$ tokens on average.

\section{Conclusion}
We proposed Long Range Arena (LRA), a new benchmark for evaluating progress on efficient Transformer research. Our new benchmark is challenging and probes at model capabilities in dealing with diverse data types and structures such as text, mathematics, and visual data. Our benchmark comprises of tasks ranging from $1K$ to $16K$ tokens. For the first time, we conduct an extensive side-by-side comparison of ten recently proposed efficient Transformer models. The experimental results show that these tasks are very challenging even for long-range Transformer models. The overall results show that there is no one-size-fits-all solution and trade-offs have to be made in terms of model quality and speed/memory. We plan to open source our code and benchmarks to facilitate future benchmarking, research and model development.

\section{Acknowledgements}
We would like to thank the following colleagues: Krzysztof Choromanski, Richard Song, Tamas Sarlos for recommendations on Performer setups. David Dohan and Manzil Zaheer for help on the BigBird implementation. Anselm Levskaya for some useful reference code for Reformers. Orhan Firat for helpful pointers. Jiaxi Tang, Jai Gupta, Zhen Qin, Che Zheng, Zhe Zhao, Da-Cheng Juan, Thomas Unterthiner, Marc Najork, Aurko Roy, Kevin Murphy, Ashish Vaswani, Niki Parmar, Mohammad Taghi Saffar, Noah Fiedel and Peter J Liu, for general feedback and discussions. We would also like to thank Drew Linsley, who provided us with help and information for setting up the path-finder benchmark.   

\bibliography{ref}
\bibliographystyle{iclr2021_conference}
\newpage
\appendix
\section{Appendix}

\subsection{LRA tasks}
This section describes the details and hyperparameters of each task. We also plan to release the configuration files along with the implementation of the models and benchmarks, that can be used to reproduce the results reported in the paper.

\subsubsection{ListOps}
Following the generation steps in \citep{nangia2018listops}, we generate our own long version of this task. We use a sequence length of $2k$ for this task. All our xformer models have an embedding dimension of $512$, $8$ heads, $6$ layers and a feed-forward dimensions of $2048$. We train all models for $5K$ steps. The [CLS] token is used and mapped into a 10 class Softmax layer for classification. 

\subsubsection{Byte-level Document Classification} We use the IMDb reviews dataset \citep{maas-EtAl:2011:ACL-HLT2011} and a sequence length of $\{1K,2K,3K,4K\}$ tokens for all models. We pick the best results across these four sequence lengths. We use a [cls] token for prediction. All the [cls] tokens from xformer encoders are passed into a two layered MLP with ReLU activations. The MLP emits a $2$-class logits for binary classification. We optimize the softmax cross entropy loss function. All xformer models are parameterized by the same number of layers, heads and hidden dimensions, namely $8$ heads, $512$ hidden dimensions and $d=2048$ for positional FFN layers. We use $6$ layers for all xformers. The learning rate is $0.05$ with weight decay of $0.1$. We use Adam with warmup. All models are trained for $20K$ steps and a batch size of $32$.

\subsubsection{Byte-level Document Matching} We use the ACL anthology network for a related article matching task. We use a sequence length of $4K$ per document ($8K$ tokens in total for two sequences). The two encoders share parameters. Similar to document classification, we use the [cls] token from xformer encoders. Let $X_1$ be the [cls] token embedding from document 1 and $X_2$ be the [cls] token embedding from document 2, the final score is computed via:
\begin{align}
Y = \text{MLP}([X_1, X_2, X_1 * X_2, X_1 - X_2])
\end{align}
where MLP(.) is a two layered MLP with relu activation functions. In lieu of the much longer sequence length, we use a batch size of $32$, embedding dimension of $128$, $4$ heads, a FFN dimension of $512$ and $4$ layers. Model is trained with Adam for $5K$ steps with a learning rate of $0.5$.

\subsection{Image Classification}
We use the gray-scaled (single channel) CIFAR10 as the image classification dataset, with 10 classes. The resolution of input images is $32 \times 32$ and after flattening the input images, we feed our xformer encoders with a sequence of $1024$ pixels. Similar to our other classification tasks, there is a classifier head on top of the xformer encoder, consisting of a two-layer  MLP with ReLU activation.  Softmax cross-entropy has been used for optimizing the parameters of the models. We trained our models for $200$ epochs and have done extensive sweeps over different hyper-parameters and found the following values leading to the best average performance across all xformers: $3$ layers, $4$ heads, $128$ as the hidden dimensions of FFN blocks, $64$ as the query/key/value hidden dimensions, and finally the learning rate of $0.01$.

\subsubsection{Generalization Gap}
For the image classification benchmark, in Section~\ref{sec:results}, we mentioned that most of the models struggle generalizing to the test set. Table~\ref{tab:image_classification} presents the train and test accuracy for different models and for almost all these models, the gap between the two scores is considerably high.    
\begin{table}[h!]
    \centering
    \begin{tabular}{l|cccccc|c|}
    \toprule
      Model &  test accuracy  & train accuracy \\
    \midrule
        Transformer &  42.44 & 69.45 \\
        Local Attention & 41.46 & 63.19 \\
        Sparse Trans. & \textbf{44.24} &  66.74\\
        Longformer& 42.22 & 71.65\\
        Linformer & 38.56 &  97.23\\
        Reformer & 38.07 &  68.45 & \\
        Sinkhorn Trans. & 41.23 & 69.21\\
        Synthesizer & 41.61 & \textbf{97.31}\\
        BigBird  & 40.83 & 71.49\\
        Linear Trans. & 42.34 & 65.61\\ 
        Performer  & 42.77  & 73.90 \\
     \bottomrule
    \end{tabular}
    \caption{Test and train accuracy of different models on Image Classification task.}
    \label{tab:image_classification}
\end{table}

While this task can be simple to solve for convectional models (e.g., accuracy of wide-resnet on gray-scale CIFAR10 with no data augmentation is $89.21$) it is rather difficult for Transformer-based models with this setup. 
Naturally, one can find ways to improve the performance with a different setup. For instance, in our setup, models are not informed about the oridinality of pixel intensities and consume them as independent symbols. We observed that learning embedding that reflects this property is rather hard for most of these models (Figure~).  If we simply replace the embedding layer with a CNN stem, we see imitate boost in the performance (e.g. replacing the embedding layer of a vanilla Transformer with a convectional stem, with $3 \times 3$ kernel, we  get accuracy of $75.32$). 

Another modification that can lead to better performance is to incorporate spatial representation that are translation invariant in Transformer models (e.g., adding 2D relative positional embedding to a vanilla transformer, we get accuracy of $61.72$). However, adding these sorts of changes make the setup digress from the original point of this task in our benchmark.

\subsubsection{Visualizations of leaned embedding by a vanilla Transformer}
\begin{figure}[h!]
    \centering
    \includegraphics[width=0.48\textwidth]{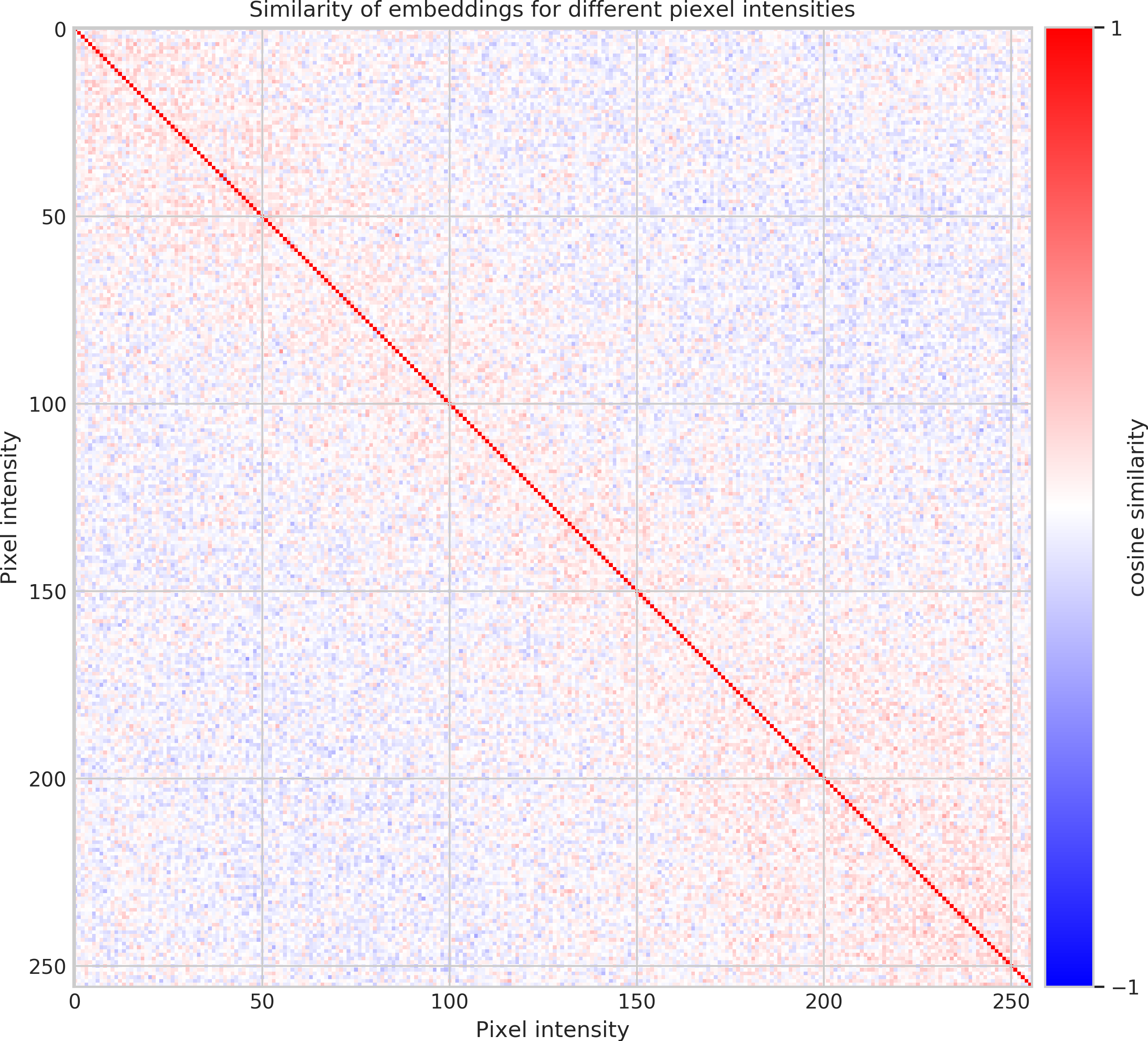}
    \hspace{4pt}
    \includegraphics[width=0.48\textwidth]{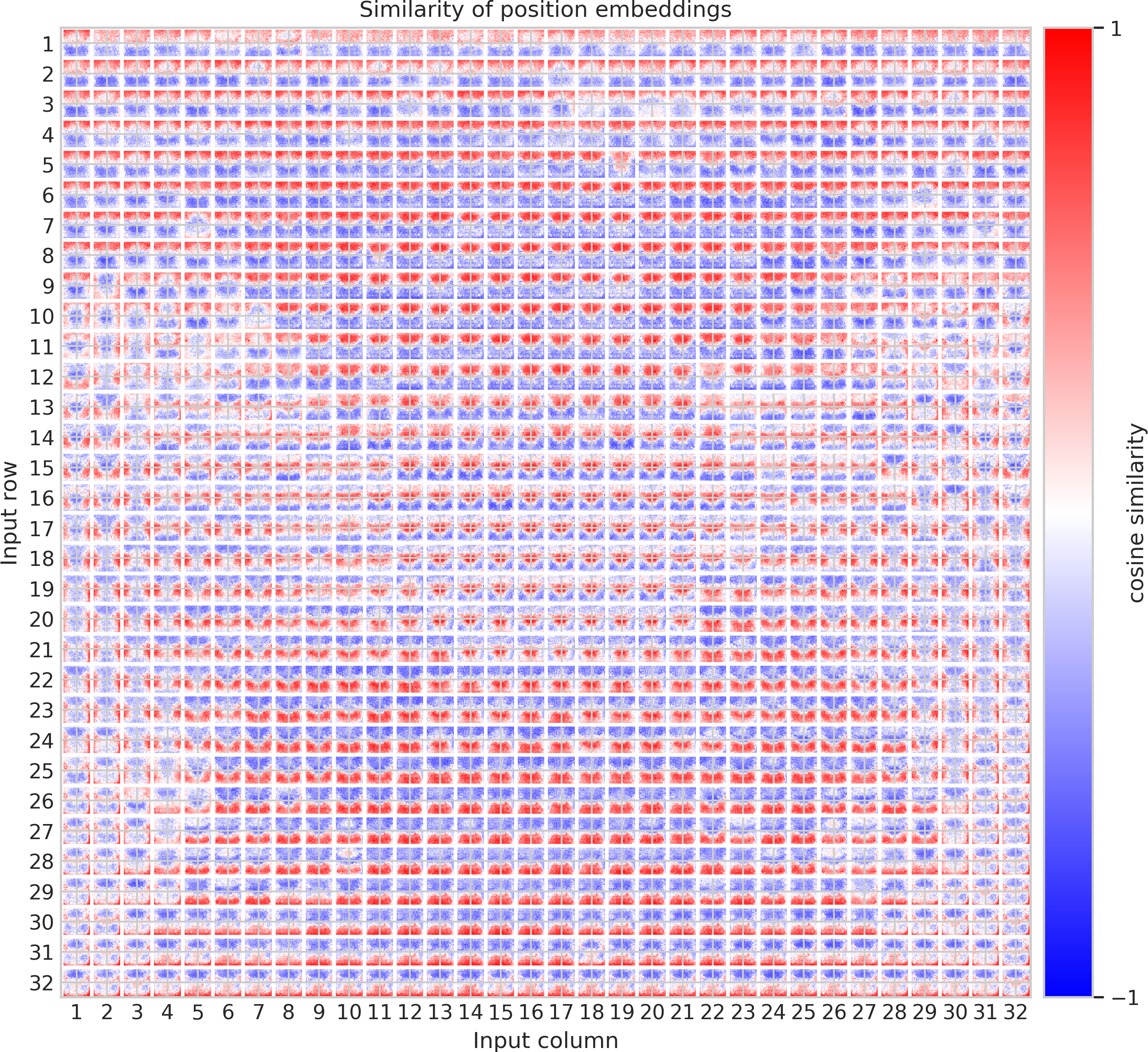}
    \caption{
        \textbf{Left:} The cosine similarity between the embedding learned for each pixel intensity. 
        \textbf{Right:} Each tile shows the cosine similarity between the position embedding of the pixel with the indicated row and column and the position embeddings of all other pixels.}
    \label{fig:cifar10_visualization}
\end{figure}

Figure~\ref{fig:cifar10_visualization} presents visualizations for the pixel intensity and positional embedding that a vanilla transformer model learns for the image classification task, on the gray-scaled CIFAR10 detest. 

On the left, we can see the pairwise similarity of learned embeddings for pixel intensities.  Although there is a higher similarity for close pixel values, the patterns from these learned embeddings do not perfectly reflect the ordinality of the pixel intensities. 
On the right, we can see the pairwise similarity of positional embeddings for different input positions. We can see that the lower the distance between two pixels is, the more similar are their learned positional embeddings. However, the spatial closeness in $y$ axis is more preserved in the learned embedding than the distances in the $x$ axis.

\subsection{Pathfinder}
Pathfinder task probes the ability of models to detect long range spatial dependencies between input features. To solve the task, a model requires to identify the target contour and trace it from one end to the other. Although Pathfinder is visually a simple task, it has been show that the clutter and variations in path shape makes the task difficult for CNN models~\citep{linsley2018learning, Kim2020Disentangling}. 

The Pathfinder task is a binary classification task and the resolution of input images is $32\times32$. Similar to image classification task, we feed our xformer encoders with a sequence of $1024$ pixels after flattening the input images. The classifier head on top of the xformer encoder is also a two-layer  MLP with ReLU activation and we use Softmax cross-entropy loss for the optimization.  We trained our models for $200$ epochs. The hyper-parameters used for the xformer model are as follow: $4$ layers, $8$ heads, $128$ as the hidden dimensions of FFN blocks, $128$ as the query/key/value hidden dimensions, and the learning rate of $0.01$.

\subsubsection{Visualization of the Attention Maps from a Vanilla Transformer}
\begin{figure}[h!]
    \centering
    \includegraphics[width=\textwidth]{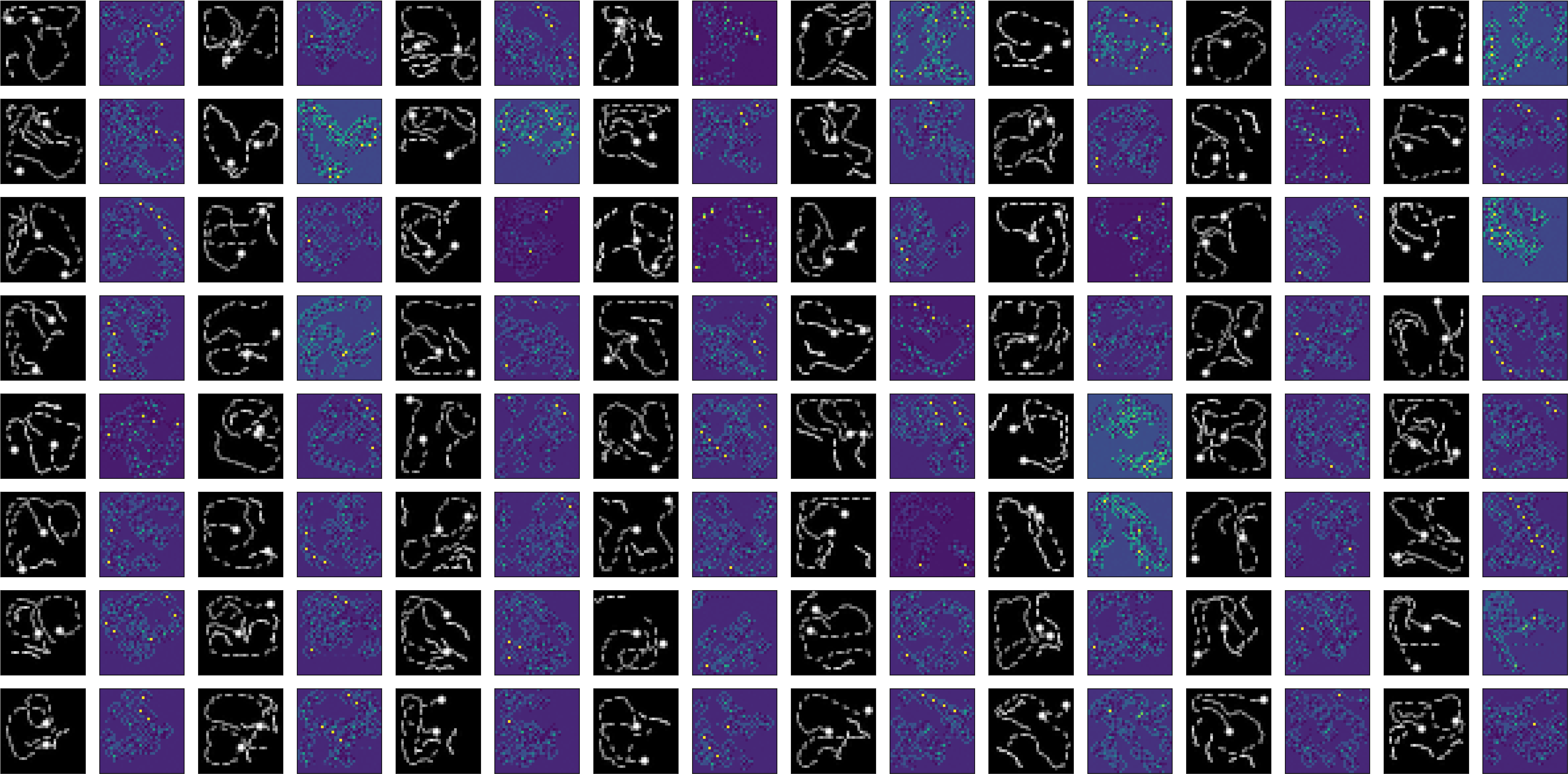}
    \caption{Attention map for different examples from the Pathfinder task. Each map presents the attention distribution, given the CLS token at the final layer as the query, averaged across all heads in a vanilla Transformer model. Note that for visualization, we use attention-rollout~\citep{abnar-zuidema-2020-quantifying} for more precise input attribution.}
    \label{fig:pathfinder_attmap_visualization}
\end{figure}

Given that transformers have many units with global receptive field, they have better potential for solving the task, compared to models with local receptive fields. Figure~\ref{fig:pathfinder_attmap_visualization} shows the attention distributions for a set of examples given on token (CLS token) as the query. We can see that the attention module collects information from different positions in input to be able to trace the target path.

We have also included a Pathfinder-X in LRA, which is similar to Pathfinder, but inputs are in higher resolutions, i.e. longer input sequences. On Pathfinder-X, we have tried two setups for training our models, first training models from scratch, second evaluating models that are trained on Pathfinder. In both cases, we found out none of the models are able to deal with/generalize to 16K input length.  

\section{Models and Implementation}
This section describes the details of our implementation. The code is primarily written in JAX and FLAX. In this section, we note specific details about certain implementations of models. We plan to release hyperparameters in a form of readme or script later.

\subsection{A Brief Overview of Model Implementations}
While most of the fine-grained details is planned to be available in the released code, we provide a brief overview of some settings of the xformer models being evaluated. For local attention, we do not use overlapping blocks. For local attention within Sinkhorn Transformer blocks, we do not overlap windows either. For Linformers, the projections are shared between key and values but not across multiple layers. For Performer models, our implementation uses FAVOR+, the more recent version in the paper \cite{choromanski2020rethinking}. 

\subsection{Special Cases of our Implementation}
This section describes several special cases in our implementation details. The diverse suite of Transformers come with a plethora of hardware constraints and implementation details. To succeed, a Transformer model needs to also `win' the hardware lottery \citep{hooker2020hardware}, i.e., having readily supported ops, kernels or accelerator support to take advantage of its technical design. This section discusses some of the trade-offs and edge cases that make comparison of several models challenging. In the end, we argue that simplicity is a virtue and not requiring any special support is a positive thing for an efficient Transformer model.
\paragraph{On CUDA kernels} 
CUDA kernels are cumbersome and are specific to GPU hardware, making it difficult to implement or use on TPU pods. Generally, these are considered to be undesirable and inconvenient in practical applications. Hence, Sparse Transformer and Longformer are implemented with \textbf{equivalent} implementations to emulate for performance. This is by applying an equivalent mask. For this reason, we do not benchmark Sparse Transformer and Longformer for speed.
\paragraph{Reformer's Implementation}
Having optimized ops to support many of Reformer's functionality is crucial. Hence, Reformer is implemented slightly differently from other Transformer models. Instead of computing tensors with batch size dimensions $B$ and head dimensions $H$, (i.e., $B \times H \times N \times d$), we compute the attention function for tensors of $N \times d$ dimensions. After which, we parallelize this function via \textsc{vmap} over the batch and head dimensions.

\section{Recommendations for Fair Comparison}
We welcome re-evaluation of our models on any task. However, we consider some hyperparameters to be \textbf{immutable} to ensure fair comparison with all models. In the case of proposing new models, the LRA table in the paper can be copied as it is as long as (1) the model size remains unchanged, (2) no pretraining is conducted, (3) no alterations to the fundamental setups (e.g., changing char level to word level or adding spatial information to the image task). We will provide more details at \url{https://github.com/google-research/long-range-arena}.

\end{document}